\documentclass{article}

\usepackage{microtype}
\usepackage{graphicx}
\usepackage{subfigure}
\usepackage{booktabs} % for professional tables
\usepackage{multirow}

\usepackage{hyperref}

\usepackage[accepted]{arxiv}

\usepackage{amsmath}
\usepackage{amssymb}
\usepackage{mathtools}
\usepackage{amsthm}
\usepackage{svg}
\usepackage{enumitem}
\usepackage[capitalize,noabbrev]{cleveref}
\usepackage{bbding}
\newlist{senum}{enumerate}{1} % Define "senum" for S1, S2, S3
\setlist[senum,1]{label=(S\arabic*)}

\newlist{penum}{enumerate}{1} % Define "senum" for S1, S2, S3
\setlist[penum,1]{label=(P\arabic*)}

\newlist{wenum}{enumerate}{1} % Define "senum" for S1, S2, S3
\setlist[wenum,1]{label=(W\arabic*)}

\newlist{eenum}{enumerate}{1} % Define "eenum" for E1, E2, E3
\setlist[eenum,1]{label=(E\arabic*)}

\newlist{rqenum}{enumerate}{1} % Define "eenum" for E1, E2, E3
\setlist[rqenum,1]{label=(RQ\arabic*)}

\theoremstyle{plain}

\theoremstyle{definition}

\theoremstyle{remark}

\usepackage[textsize=tiny]{todonotes}
\usepackage{tcolorbox}
\tcbuselibrary{listingsutf8}

\newtcolorbox[auto counter, number within=section]{definitionbox}[2][]{%
    colback=gray!5!white,  % Background color
    colframe=gray!75!black, % Border color
    fonttitle=\bfseries,   % Title font
    coltitle=black,        % Title text color
    colbacktitle=gray!40!white, % Title background color
    title=Definition~\thetcbcounter: #2, % Title text
    #1                      % Optional parameters
}

\icmltitlerunning{Position: Episodic Memory is the Missing Piece for Long-Term LLM Agents}

\begin{document}

\twocolumn[
\icmltitle{Position: Episodic Memory is the Missing Piece for Long-Term LLM Agents}



\icmlsetsymbol{equal}{*}

\begin{icmlauthorlist}
\icmlauthor{Mathis Pink} {yyy}
\icmlauthor{Qinyuan Wu} {yyy}
\icmlauthor{Vy Ai Vo} {comp}
\icmlauthor{Javier Turek} {ea}
\icmlauthor{Jianing Mu} {sch}
\icmlauthor{Alexander Huth} {sch}
\icmlauthor{Mariya Toneva} {yyy}
\end{icmlauthorlist}

\icmlaffiliation{yyy}{Max Planck Institute for Software Systems}
\icmlaffiliation{comp}{  }
\icmlaffiliation{ea}{EarthDynamics.ai}
\icmlaffiliation{sch}{University of Texas at Austin}

\icmlcorrespondingauthor{Mathis Pink}{mpink@mpi-sws.org}

\icmlkeywords{Machine Learning, ICML}

\vskip 0.3in
]



\printAffiliationsAndNotice{\icmlEqualContribution} % otherwise use the standard text.

\newcommand{\mariya}[1]{{\color{purple} #1 [MT]}}
\newcommand{\qinyuan}[1]{{\color{blue} #1 [QW]}}
\newcommand{\mathis}[1]{{\color{magenta} #1 [MP]}}
\newcommand{\vy}[1]{{\color{red} #1 [VV]}}
\newcommand{\vvcomment}[1]{{\color{red} \textit{[VV] #1}}}
\newcommand{\javier}[1]{{\color{brown} #1 [JT]}}
\newcommand{\jianing}[1]{{\color{teal} #1 [JM]}}

\begin{abstract}

As Large Language Models (LLMs) evolve from text-completion tools into fully fledged agents operating in dynamic environments, they must address the challenge of continuous learning and long-term knowledge retention. Many biological systems solve these challenges with episodic memory, which supports single-shot learning of instance-specific contexts. Inspired by this, we present a framework for LLM agents, centered around five key properties of episodic memory that underlie adaptive and context-sensitive behavior. With various research efforts already covering portions of these properties, this position paper argues that now is the right time for an explicit, integrated focus on episodic memory to catalyze the development of long-term agents. To this end, we outline a roadmap that unites several research directions under the goal to support all five properties of episodic memory for more efficient long-term LLM agents.

\end{abstract}

\section{Introduction}
\label{introduction}

Large Language Models (LLMs) are rapidly expanding beyond their origins as text-completion engines. Instead, they are evolving into agentic systems capable of taking meaningful actions in complex environments~\citep{xi2023risepotentiallargelanguage}. This transformation can enable a range of real-world applications, including autonomous research assistance~\citep{schmidgall2025agent}, aiding in literature reviews, data analysis, and hypothesis generation; personalized customer support~\citep{li2024personalllmagentsinsights}, where they can recall prior interactions to provide consistent and tailored assistance; and interactive tutoring systems~\citep{lin2023artificial}, which track learning progress, and revisit challenging concepts to ensure effective and personalized education. 
These diverse applications hint at a vast potential of LLMs to enable intelligent agents capable of meaningful and context-sensitive interaction.

Operating and reasoning over extended timescales in dynamic interactive contexts demands that an agent not only recalls what happened, but also when, how, why, and involving whom. Such rich traces of past events, motivations, and outcomes form the basis of context-sensitive behavior—especially crucial in large-scale projects involving human stakeholders and multiple actors. For example, a future long-term LLM agent that is supposed to assist in the ongoing development of a massive software project such as Linux—which has spanned decades, encompasses over 40 million lines of code, and additionally involves countless past contributions, issues, comments, notes, and feature requests—would need to continuously integrate and reason about a vast, evolving historical context while adapting to new requirements. 
Core necessities for this kind of system are constant computational cost per new token and a stable or improving performance over time. 

Ongoing research directions attack the problem of long-term retention and adaptation from different angles and have made impressive progress.
However, we are still lacking approaches that maintain relevant contextualized information over long time frames at a constant cost without degrading performance---necessities for a widespread adoption of LLM agents in many long-term settings.

Meanwhile, many biological systems solve the demands for acting in a continually evolving environment with a dedicated memory system that allows for both fast and slow learning: episodic memory \citep{McClelland1995, Schwartz2001, OReilly2002, OReilly2014,Kumaran2016,Liao2024}. \textbf{In this position paper, we argue that the growing demand for LLM agents to operate effectively over extended timescales, alongside ongoing advances in long-context models, external memory systems, and efficient fine-tuning methods, makes episodic memory a timely framework to unify efforts for enabling truly long-term LLM agents.} %While there have been calls for episodic memory in AI systems before, \textbf{in this position paper, we argue that the convergence of two rapid developments—emerging applications for long-term LLM agents and advances in different areas of long-term LLM memory--spanning context expansion, external memory, and efficient finetuning--makes episodic memory a timely unifying goal for enhancing LLM agents' reliability and efficiency.}

\begin{figure*}[t]
    \centering
    \includegraphics[width=0.78\textwidth]{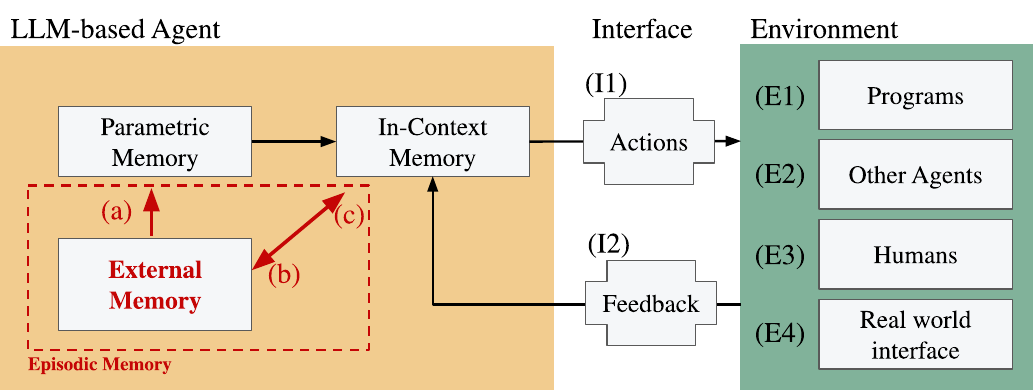}
    \caption{LLM-Agents with an Episodic Memory system. The LLM agent acts on and gets feedback from an environment. Feedback can come in the form of outputs from programs (E1), from other agents (E2), humans (E3), as well as external real-world data (E4). Actions can modify parts of the environment, and provide feedback for humans or other agents in the environment. Within the agent, an external memory system acts as a bridge between parametric and in-context memory while allowing for fast \emph{encoding} of and \emph{retrieval} into in-context memory (the LLM's context window). (a) \emph{Consolidation}: Episodes in the external memory are consolidated into a model's broader parametric memory to avoid capacity limitations and allow for generalization to new semantic knowledge and procedural skills based on specific instances. (b) \emph{Encoding}: Limited in-context memory can offload its content into external memory. (c) \emph{Retrieval}: Stored episodes can later be retrieved and used to reinstate representations into in-context memory.}
    \label{fig:episodic-memory-llm-agent}
\end{figure*}

To lay out the argument for this position, we proceed as follows: In Section \ref{sec:em_properties}, we operationalize the concept of episodic memory for LLM agents by highlighting five key properties that distinguish it from other biological types of memory that are also desirable for LLM agents. We proceed to argue in Section \ref{sec:previous-work} for episodic memory as a unifying goal by showing how various existing approaches to improve LLM memory target different properties that are united in episodic memory. In Section \ref{sec:unifying}, we highlight how unifying these threads under a common goal can spur more holistic progress and outline a roadmap toward implementing episodic memory. % including new benchmarks, the addition of contextual detail to RAGs, practical strategies for distilling cached states into the broad reservoir of parametric knowledge, as well as the idea of making use of advancements in long-context attention to guide new, more promising architectures. 
Lastly, in Section \ref{sec:alternatives}, we discuss alternative views under which episodic memory would not be necessary for long-term LLM agents.

\section{Operationalizing EM for LLMs}
\label{sec:em_properties}

To transfer the concept of episodic memory from cognitive science to the context of LLM agents, we highlight five properties of episodic memory that are useful for LLM agents, and that distinguish episodic memory from other memory types in animals and humans. These five properties naturally cluster into two categories: properties that concern the way that the system operates with the memory---namely, long-term storage, explicit reasoning, and single-shot learning, and properties that concern the content of the stored memory---namely, instance-specific and contextualized memories. We first discuss how the combination of these five properties distinguishes episodic memory from other types of memories in animals and humans, and then detail each property and its utility for LLM agents.

\label{taxonomy}

\subsection{Unique Combination of EM Properties}

Episodic memory is one of multiple memory systems that exist in animals and humans, distinguished by its unique combination of properties (Table \ref{memory_table_biology}). Other biological memory systems that share some, but not all, properties of episodic memory are 1) procedural memory \citep{Milner1962,CohenSquire1980}, which allows for long-term storage of memories for implicit operations or task behaviors, such as producing a sequence of a particular type, rather than reasoning about the sequence; 2) semantic memory \citep{CollinsQuillian1969, Tulving1972}, which allows for long-term storage of factual knowledge and explicit reasoning with these stored memories, but lacks specificity to single instances of acquired information and its context; and 3) working memory \citep{Baddeley1986, BaddeleyHitch1974}, which can share many of the highlighted properties of episodic memory except for the important fact that it does not allow for long-term storage. The unique combination of important properties in episodic memory makes it a promising candidate for translation to AI systems.

\subsection{Importance of EM Properties for LLM Agents}

\begin{table}[b]
\caption{Properties of episodic memory in comparison to other relevant forms of memory in animals and humans.}
\label{memory_table_biology}
\centering
\small % Reduce font size
\begin{tabular}{lp{0.5cm}p{0.8cm}p{0.6cm}p{0.9cm}p{1.2cm}}
\toprule
%\textbf{Memory Type} & \textbf{Instance-specificity} & \textbf{Long-term} & \textbf{Single-shot} & \textbf{Explicit} & \textbf{Contextual \ relations}\\
\textbf{Memory Type} & \textbf{Long-term} & \textbf{Explicit} & \textbf{Single-shot} & \textbf{Instance-specific} & \textbf{Contextual \ relations}\\
\midrule
Episodic & \Checkmark & \Checkmark & \Checkmark & \Checkmark & \Checkmark \\
Procedural & \Checkmark &  \(\times\) & \(\times\) & \(\times\) & \(\times\) \\
Semantic & \Checkmark & \Checkmark & \(\times\) & \(\times\) & \(\times\) \\
Working & \(\times\) & \Checkmark & \Checkmark & \Checkmark & \Checkmark \\
\bottomrule
\end{tabular}
\end{table}

\subsubsection{Episodic Memory Operations}

\textbf{Long-term storage.}
In humans and other animals, episodic memory functions as a form of long-term memory, capable of storing knowledge throughout an individual’s lifetime \citep{Conway2001, Mayes2001, Squire1996, Hampton2004}. This distinguishes it from working memory, which is transient. %For LLM agents, an effective episodic memory system must similarly support the retrieval of stored information regardless of the number of tokens that are observed in between the point at which the memory was first encoded and the point at which the memory and its context become relevant for recall. 
For LLM agents, an effective episodic memory system must similarly support memory retrieval across any number of tokens. This requires mechanisms for long-term memory that maintain an agent's performance throughout a continual interaction with an environment. %Instead of degrading with more experience, a
An adaptive long-term agent should not only prevent a degradation in performance over time-it should also be able to improve by learning new general knowledge and skills.

\textbf{Explicit reasoning.}
In classical theories of human memory, episodic memory is described as a subset of declarative or explicit memory \citep{Squire1996, Hampton2004}. A defining feature of explicit memory is the ability to reflect and reason about the memory content. In the context of LLM agents, the explicitness of memory is necessary as agents need to be able to answer direct queries about stored information or use this information in explicit internal reasoning processes.

\textbf{Single-shot learning.}
A key characteristic of episodic memory, as emphasized in complementary learning systems theory, is its ability to be acquired based on a single exposure \citep{Liao2024, Schwartz2001, OReilly2002, OReilly2014, McClelland1995, Kumaran2016, Das2024}. This fast learning enables the rapid encoding of unique experiences or events. For LLM agents, this capability is particularly crucial in environments where continual deployment may not provide multiple variations or repetitions of specific events. Certain occurrences in an environment may happen only once, necessitating an episodic memory system that is capable of effectively capturing and utilizing information from single exposures.

\subsubsection{Episodic Memory Content}

\textbf{Instance-specific memories.}
Episodic memory stores information specific to an individual sequence of events along with their distinct temporal contexts \citep{Sugar2019, Colgin2008}. This specificity allows episodic memory to capture details unique to a particular occurrence, enabling its application in agentic environments where reasoning about specific past actions and their consequences matters. This can include past lines of reasoning that were associated with a decision to be made by an LLM agent. 

\textbf{Contextual memories.} Episodic memory binds context to its memory content, such as when, where, and why an event was encountered \citep{Eichenbaum2014, OKeefe1978, Eichenbaum2015}. The ability to store many contextual relations associated with a specific event enables retrieval based on contextual cues as well as explicit recall of context. For LLM agents, this property is important to not only remember that a specific event happened in the past, but also when, why, and in which broader context it happened.

\section{Current Approaches} 
\label{sec:previous-work}

While many methods currently exist to modify and augment LLM memory, we argue that they fall short of the memory properties that would enable effective, long-term LLM agents. We group existing methods that seek to improve the memory of LLMs into three categories that are relevant to episodic memory: 

\vspace{-0.3cm}
\begin{enumerate}
    %\vspace{-0.1cm}
    \item \textbf{In-Context Memory} methods extend the effective context length by optimizing computational efficiency and length generalization;
    \vspace{-0.1cm}
    \item \textbf{External Memory} methods augment a model's in-context memory capacity with a separate module, often with reduced GPU memory requirements and/or computational cost;
    \vspace{-0.1cm}
    \item \textbf{Parametric Memory} methods modify the LLM parameters that encode memories (primarily learned from the language modeling training data).
\end{enumerate}
\vspace{-0.3cm}

In this section, we discuss examples in each category that capture different properties of episodic memory (Table \ref{memory_table}). More importantly, we highlight their shortcomings in supporting episodic memory for LLM agents in isolation.

\begin{table}[t]
\centering
\caption{
%Memory-related m
Methods for in-context, external, and parametric memory do not cover all features of episodic memory. $\sim$ is used for cases where it is unclear whether an aspect of episodic memory is properly satisfied by a method.}
\label{memory_table}
\resizebox{\columnwidth}{!}{%
\begin{tabular}{l l c c c c c}
\toprule
\multicolumn{2}{l}{\textbf{Memory Approach}} & 
\rotatebox{90}{Long-term} & 
\rotatebox{90}{Explicit} & 
\rotatebox{90}{Single-shot} & 
\rotatebox{90}{Inst.-specific} & 
\rotatebox{90}{Contextual rel.} \\

% & Test-Time-Training  &  \(\times\)  &   \Checkmark  &   \Checkmark  &  \Checkmark   &   \Checkmark  \\
\midrule
\multirow{2}{*}{\textbf{In-Context}}
 & KV-Compression           & \(\times\)   &  \Checkmark  & \Checkmark   & \Checkmark   &  \Checkmark  \\
 & State-space-model   &  \(\times\)  &  \Checkmark   &  \Checkmark  &  \Checkmark  &  \Checkmark   \\
\midrule

\multirow{2}{*}{\textbf{External}}
 & RAG                 &   \Checkmark  &   \Checkmark  &  $\sim$   &  $\sim$   & \(\times\)  \\ 
 & GraphRAG            &    \Checkmark &   \Checkmark  &   $\sim$  &  $\sim$   & $\sim$   \\
\midrule

\multirow{3}{*}{\textbf{Parametric}}
 & Efficient Fine-tuning &  \Checkmark   &  \Checkmark  &  $\times$  &  $\times$  &  $\times$  \\
 & Knowledge Editing   &   \Checkmark   &  $\sim$  &  $\times$  &  $\sim$  &  $\times$  \\
 & Context Distillation &  \Checkmark   &  \Checkmark  &  $\times$  &  \Checkmark  &  $\times$  \\ 
%\midrule

\bottomrule
\end{tabular}%
}
\end{table}

\subsection{In-Context Memory}

In-context memory (ICM) allows LLMs to perform single-shot, instance-specific, and contextualized learning by enabling them to directly attend to representations of encountered sequences (Table~\ref{memory_table}). ICM capacity is either tightly limited or extensible but expensive to scale, often requiring sequence parallelization. Recent works seek to extend it by increasing the context window, but models struggle with length generalization beyond training exposures. We review existing methods and their limitations in addressing these challenges.   

One active research direction focuses on extending the in-context window to handle significantly longer sequences, enabling LLMs to perform reasoning over extended contexts. This advancement brings LLMs closer to mimicking episodic memory, as it allows models to retain and utilize information across longer contexts. However, transformer-based LLMs face significant challenges, including the high computational cost of processing long sequences and limitations in length generalization. 
Recent research has sought to address these challenges by reducing memory usage, optimizing inference time, and improving long-sequence generation. Despite these advancements, current methods have yet to achieve robust, persistent memory capabilities necessary for long-term, open-ended, and context-aware reasoning. Below, we briefly review existing methods and their limitations.

\textbf{Memory reduction.}
For transformer-based LLMs, several methods aim to reduce memory and computation costs. 

Sparsification and compression methods selectively retain relevant information to optimize memory usage. Sparsification strategies optimize memory by restricting attention computations to the most relevant parts of the sequence~\citep{lou2024sparser}, reducing both storage and computational overhead. Similarly, forgetting mechanisms remove less useful tokens to maintain efficiency~\citep{anonymous2024forgetting}. Other compression-based approaches dynamically reduce KV cache size by storing only the most important tokens and key-value pairs~\citep{Liu2023scissorhands, ge2024modeltellsdiscardadaptive, tang2024razorattentionefficientkvcache}. Adaptive strategies further refine compression across layers~\citep{yang2024pyramidinferpyramidkvcache, nawrot2024dynamicmemorycompressionretrofitting} or merge similar states to minimize redundancy~\citep{liu2024minicachekvcachecompression}.

Quantization methods reduce memory footprints by lowering precision or selectively storing information. Quantization techniques store key-value pairs at reduced precision~\citep{liu2024kivi, hooper2024kvquant10millioncontext, yue2024wkvquantquantizingweightkeyvalue, duanmu2024skvqslidingwindowkeyvalue}, allowing for larger context windows with a relatively minimal performance degradation. 

\textbf{Inference time reduction.}
Efficiency improvements during inference focus on optimizing KV cache management and parallelization. Techniques such as paged caching~\citep{Kwon2023, lee2024infinigenefficientgenerativeinference, zheng2024sglang} dynamically allocate memory to accommodate longer sequences without excessive overhead. Other methods leverage GPU memory pooling and adaptive chunking~\citep{lin2024infinitellmefficientllmservice, agrawal2024mnemosyneparallelizationstrategiesefficiently} to process extended contexts efficiently while maintaining fast retrieval and computation speeds.  
Other strategies improve efficiency by reusing KV tensors across layers~\citep{ye2024chunkattentionefficientselfattentionprefixaware, brandon2024reducing}

Recent work has aimed to introduce episodic memory in LLMs by structuring token sequences into retrievable events~\citep{fountas2024humanlikeepisodicmemoryinfinite}, enhancing long-context reasoning and outperforming retrieval-based models. However, a fundamental challenge remain the increasing memory and retrieval costs: maintaining the full KV-Cache for an entire interaction history can quickly become impractical, especially in large-scale, long-duration, and multimodal applications. This limitation is inherent to KV-Cache management systems, which must retain the entire cache, leading to significant storage and computational overhead.

\textbf{Transformer alternatives to reduce both memory and inference time.} 
In addition to optimizing KV cache storage and management, alternative architectures have been proposed to address the limitations of standard transformers in both memory and inference time.

Linear attention~\citep{li2020linear, katharopoulos2020transformers} approximates full self-attention using kernel-based or low-rank transformations, significantly reducing computational complexity and improving efficiency for long-sequence processing. 
State-space models (SSMs)~\citep{peng2023rwkvreinventingrnnstransformer, gu2023mamba} further achieve linear scaling for sequence handling by maintaining a fixed-size representation, making them inherently memory-efficient. Hybrid architectures~\citep{goldstein2024goldfinchhighperformancerwkvtransformer} combine these techniques with transformers to compress KV-cache sizes while preserving strong performance.  
Other alternatives restructure the transformer architecture itself to enhance efficiency. Some models~\citep{sun2024cacheoncedecoderdecoderarchitectures} modify the decoder structure to reduce memory usage and latency, while others~\citep{pang2024anchorbasedlargelanguagemodels} compress sequence information into compact representations to improve inference speed and scalability.  

These methods enhance ICM efficiency, but their reliance on compression, approximation, and selective retention comes with limited support for long-term reasoning with episodic memory. This limitation highlights the need for an external memory structure that retains past information. Methods of KV-cache optimization can also discard older context, leading to irreversible information loss and different model behavior~\citep{kirsten2024impact}. 
Generally, methods with a constant cost, like SSMs, struggle to handle a continually expanding interaction history in dynamic environments, while methods with an increasing state representation increase in both inference time and GPU memory requirements. 

\textbf{Length Generalization.}
Length generalization refers to a model's ability to maintain understanding over long sequences, preventing degradation of performance such as forgetting or losing context midway through processing~\citep{liu2024lost}. In essence, humans avoid SSMs' trade-offs by storing compressed representations and retrieving knowledge adaptively, allowing us to manage expanding information effortlessly.

To address this, lightweight solutions~\citep{yen2024longcontextlanguagemodelingparallel, xiao2024infllmtrainingfreelongcontextextrapolation} create adapters to process and retrieve long inputs before passing the content to the LLM.  
 Other approaches~\citep{han-etal-2024-lm} refine attention patterns and positional encodings to enhance long-context comprehension.
 Alternative architectures~\citep{ye2024differentialtransformer,dai1901transformer} improve long-context learning through mechanisms like differential attention and segment-level recurrence. 
Another promising approach embeds test-time information into the model's parameters, creating a form of long-term memory~\citep{sun2024learninglearntesttime, behrouz2024titanslearningmemorizetest}, combining attention with neural memory modules, enabling adaptability for long contexts but at the cost of increased inference overhead. These approaches have limited capacity and still face eventual forgetting over very long sequences. %Additionally, the lack of interaction with parametric memory prevents true long-term learning, leaving models incapable of persistently retaining knowledge.

\subsection{External Memory}
\label{subsec:externalmem}

Many methods propose a separate memory module that stores information when it exceeds the effective operating span of the model. These augmented memory models are usually evaluated on tasks which require using that stored information. As such, these methods typically have long-term and explicit memory (Table \ref{memory_table}). However, they often lack information that relates the stored memories to one another---especially contextual details on how the model acquired the memory, or details to help differentiate specific instances. They are typically not evaluated for single-shot learning, especially for specific instances. And finally, there is a lack of proposals to generalize information from these instances and update parametric memory (Figure \ref{fig:episodic-memory-llm-agent}a). Below we review some relevant external memory methods and elaborate on key examples to illustrate these shortcomings.

\textbf{Slot-based memory with recurrent controllers.} A key advance in memory augmentation in the pre-transformer era was the formulation of learnable memory modules external to the main neural network \citep{bordes2015largescalesimplequestionanswering, graves2014neuralturingmachines, sukhbaatar2015endtoendmemorynetworks}. External memories were stored in individual slots and updated via a recurrent memory controller. These models were shown to retain longer-term information than vanilla long-short term memory (LSTM) networks. %More recently, 
Similar memory augmentation methods have been adapted for transformers \citep{wu2022memformer}. However, these methods lack a way to store contextual details that LLM agents would need in an episodic memory, as they strongly depend on the details available in the input data. One exception devised a method to record temporal relationships between memories \citep{graves2016hybrid}, but this has yet to be seen in augmented LLMs.

\textbf{Distributed vs. slot memory.} An issue with slot-based memory modules is that they are capacity-limited, both by the number of slots and the dimensionality of each slot representation. While these models adopt forgetting mechanisms to mitigate this, the capacity limit still affects how long memories can be stored. Another approach addresses this downside by storing external memories in a sparse, distributed fashion \citep{wu2018kanerva} instead of in slots. Recent work \citep{das2024larimarlargelanguagemodels} integrated distributed memory in an LLM, and showed that the model can recall a greater number of facts over longer contexts, compared to baseline LLMs. While they demonstrate how they can perform one-shot memory updates (fact-editing), they do not evaluate single-shot learning of novel facts.

\textbf{RAG and GraphRAG methods.} Retrieval Augmented Generation (RAG) methods maintain an external database of information that is added to the input data to augment LLM generation. Naive RAG implementations encode chunks of text using embedding models \citep{gao2023enablinglargelanguagemodels}, typically without much metadata or contextual detail about the original text. (One exception is work that preserves the order of retrieved text from the database \citep{yu2024defenserageralongcontext}.) And while text embedding models can capture some similarity relationships between embeddings, they do not encompass the rich set of relationships that LLM agents will likely need for most applications. GraphRAG models replace the vector embedding database with a structured graph that explicitly encodes relationships as connections between nodes \citep{peng2024graphretrievalaugmentedgenerationsurvey}. Still, these graphs encode a limited number of relationship types, even when researchers branch out beyond pre-existing datasets and learn to build the graphs directly from the input text \citep{li2024graphreaderbuildinggraphbasedagent, edge2024localglobalgraphrag, gutiérrez2025hipporagneurobiologicallyinspiredlongterm}. As such, they also lack rich contextual detail.

\textbf{External storage of past LLM inputs and outputs.} Another type of approach maintains a database of pasts LLM inputs to avoid recomputing predictions to similar future inputs \citep{wu2022memorizing, khandelwal2020generalizationknnlm, yogatama2021semiparametriclm}. Here, contextual information (e.g. details that differentiate specific instances) will only be stored when explicitly given in the LLM input text. That is, the memory is much more dependent on input data, limiting test-time generalization. One proposal to mitigate this formulates a long-term memory module for context that is updated with LLM activations based on the current inputs \citep{behrouz2024titanslearningmemorizetest}. Other approaches additionally store LLM outputs, such as generated text \citep{cheng2023selfmemrag}, summarizations \citep{wang2024selfcontrolledmemllm, lee2023promptllmlongchatbotmem}, chain-of-thought steps \citep{liu2023thinkinmemoryrecallingpostthinkingenable, lu2023memochattuningllmsuse}, and extracted relation triples \citep{modarressi2025memllmfinetuningllmsuse}. One approach specialized for chat interactions stores timestamps and user personality profiles as context \citep{zhong2024memorybankenhancinglargelanguage}. These modifications enable storage of contextual details useful for LLM agents. However, specifying the type of contextual detail is restrictive, so it is preferable to combine this with a more learnable and flexible mechanism for storing context.

\textbf{Learning to interact with external memory.} The approaches described above may fine-tune or instruct the LLM to interact with and update external memory. That is, the LLM learns the functions of a memory controller. For example, several RAG approaches fine-tune the LLM to make better use of the retrieved content \citep{gao2023enablinglargelanguagemodels}. Other approaches define how LLMs should interact with memory, requiring them to learn specific API calls \citep{modarressi2025memllmfinetuningllmsuse} or memory hierarchies \citep{packer2024memgptllmsoperatingsystems}. These provide possible mechanisms to add information to external memory, such as contextual details and specific instances. However, most current work does not consider how to modify the LLM to generalize across specific instances to store new knowledge in LLM parameters (Figure \ref{fig:episodic-memory-llm-agent}a). \citet{behrouz2024titanslearningmemorizetest} propose one way to generalize across instances by adding a data-independent memory system (a.k.a. meta-memory, persistent memory) in addition to a more data-dependent memory module. However, the data-independent memory is considered to be closer to task memory than knowledge distillation, and the meta-memory parameters are kept separate from the LLM itself.

\subsection{Parametric Memory}
This type of memory allows LLMs to process the information in the input to obtain well-suited output. Parametric memory values are initially learned through back-propagation with a pretraining dataset.
During this process, the parametric memory tends to capture general knowledge and rules ranging from syntax to common sense and factual knowledge.
Due to the sheer size of the parametric memory, the amount of data needed for pre-training is usually very large, following power laws \cite{kaplan2020scalinglawsneurallanguage}.
Generally, parametric memory is fixed after training, i.e., does not change with the input at inference time. 

A relevant research direction in parametric memory focuses on adapting LLM parameters to specific domains, tasks, or applications when given limited resources. Efficient fine-tuning methods have been developed in recent years to tackle the runtime and memory consumption of this process. Alternatively, distillation techniques have been proposed to update knowledge and propagate it through a model. A key challenge is the need for updating specific factual knowledge without interfering with other knowledge. Some facts may change over time, requiring surgical precision to update the parameters of a model. The line of work that proposes these updates is known as knowledge editing.

\textbf{Efficient Fine-tuning.} Various works have been proposed to reduce the computational needs (hardware memory) of updating a model to a specific domain. Among these, Low-Rank Adaptation (LoRA) \cite{hu2022lora} applies additive low-rank approximation updates to shift the model parameters.
Several methods proposed other ways to further improve efficiency by reducing and localizing updates \cite{wang2024roselorarowcolumnwisesparse,va2022DyLoRA,xu-etal-2021-raise, yin2024lofitlocalizedfinetuningllm}. Other work learned modifications on representations instead of parameters \cite{wu2024reftrepresentationfinetuninglanguage,yin2024lofitlocalizedfinetuningllm}.
In all cases, fine-tuning methods require a dataset to adapt a model for a specific task or domain, i.e. they are not capable of single-shot learning or capturing instance-specific and contextually rich information. 
On the other hand, additional fine-tuned adapter parameters are often frozen after the fine-tuning process, supporting the long term storage of information. Moreover, these methods tend to preserve the reasoning capabilities while updating the model with newly captured information \cite{wu2024reftrepresentationfinetuninglanguage}. %As many of these methods learn additional parameters, it helps reduce the catastrophic forgetting effects of fine-tuning sensible to agents and continual learning systems.  

\textbf{Knowledge Editing.} As the environment evolves over time, some factual knowledge becomes outdated (e.g., the president of a country may change after the elections). Knowledge editing methods aim to make modifications to the factual knowledge in parametric memory with targeted updates while avoiding interference with other facts. %In this area, we find various families of methods for knowledge editing. 
In ROME \citep{meng2023locatingeditingfactualassociations} and MEM-IT \citep{meng2023masseditingmemorytransformer}, the first step is to find relevant parameters (in MLPs) that influence the specific fact through causal interventions and then update the related parameters with low-rank model edits. An alternative research direction proposes to train a hyper-network \citep{decao2021editingfactualknowledgelanguage,tan2024massiveeditinglargelanguage} that predicts the amount of change for each parameter given the knowledge to be edited. A different method, SERAC, stores the set of edits in an external memory, combined with a scope detector and a counter-factual model to decide when and how to apply the edits\citep{mitchell2022memorybasedmodeleditingscale}. %A scope classifier is learned to detect inputs that have edited facts, and a counter-factual model to generate the output conditioned to the input and the relevant edit. %All the knowledge editing methods work on facts and are unable to contextualize memories. 
All knowledge editing methods work on facts which are inherently context-free, making it impossible to contextualize the edited knowledge in the history of the agent-environment interaction. However, they mimic the episodic memory traits of learning from a single instance, while enabling long-term retention.

The problem of knowledge editing has been extended to a continual learning setting, where edits are required sequentially over time to correct a model. This leads to the sequential editing problem: hyper-network prediction quality decreases because they fail to reflect the updated model, and low-rank parameter updates interfere with one another causing catastrophic forgetting~\cite{gupta2024modeleditingscaleleads}.
MELO \citep{yu2023meloenhancingmodelediting} adapts dynamic LoRA to this problem and introduces a vector database to search the selections of the blocks to be dynamically activated within the LoRA matrices for each layer. WISE \citep{wang2024wiserethinkingknowledgememory} adds duplicates of the MLP's output parameters for some layers in the network, and updates them with each new edit set. A routing mechanism decides whether to use the original layer or the updated one. It further uses sharding and merging \citep{yadav2023tiesmerging} to distribute the edits into random subspaces to improve generalization and parameter utilization. 

While continual learning-based knowledge editing allows models to integrate updates over time, it has fundamental limitations. Edited knowledge often lacks generalization, struggling with inferring new relationships or reasoning over multiple steps~\citep{berglund2023reversal,yang2024large}. This highlights a key challenge—knowledge editing methods can introduce updates but do not always ensure deeper understanding or adaptability.

\textbf{Context Distillation.} The idea behind these techniques is to transfer in-context learned information, abilities, and task-understanding by distilling them into model parameters.
\citet{snell2022learningdistillingcontext} proposed to use distillation when the teacher and the student are in the same model, but less in-context information is given to the student. This would enable the student to learn skills and express knowledge that would otherwise depend on including information and instances in costly and limited in-context memory. %without further scratchpad reasoning.
Further, \citet{distillingcontext-2} proposes to exploit context distillation to inject and propagate knowledge through a model. The original model is provided with new definitions and continuations. The distillation process updates a copy of the model with only the generated continuation, conditioning the updated model to the new entities implicitly. This helps to propagate the information into the parameters (i.e., consolidating it) and thus improving inference with such entities. 

\section{Episodic Memory as a Unifying Framework} 
\label{sec:unifying}

Although current work has advanced context-sensitive LLMs that are capable of handling longer sequences, it does not yet deliver efficient learning that could support long-term LLM agents. Existing methods—which extend in-context (working) memory, integrate external memory, or update parametric memory—only address subsets of episodic memory’s five essential properties, as discussed in Section \ref{sec:previous-work}. %Individually, t
These approaches remain fragmented, impeding the immediate assimilation of new experiences and gradual improvement over time.

We propose that enabling episodic memory offers a unifying perspective that will combine and extend existing methods to advance the capabilities of LLM agents. By incorporating long in-context memory, external memory, and mechanisms for updating parametric memory, agents can more seamlessly adapt to new information, consolidate it, and prevent escalating costs or performance degradation during extended interactions with an environment. This view is based on Complementary Learning Systems Theory \citep{OReilly2014,Kumaran2016, arani2022learningfastlearningslow}, in which episodic memory is part of a fast-learning system that stores information from individual instances. Over time, that information is consolidated into a slow-learning system that stores more stable, durable knowledge. 

In Figure \ref{fig:episodic-memory-llm-agent}, we present a general architecture and framework that combines these elements under the overarching goal of enabling all five key features of episodic memory for LLM agents as detailed in Section \ref{sec:em_properties}. As a roadmap to enable episodic memory in LLM agents, we specifically call for four main research directions (encoding, retrieval, consolidation, and benchmarks), and formulate six research questions under these areas below.

\subsection{Encoding}

RQ1: \emph{How to store information from in-context memory in a long-term external memory store?}

An external memory store is essential for retaining experience in a structured way that preserves the context of individual instances (Fig.\ref{fig:episodic-memory-llm-agent}, arrow (b)). A straightforward approach is to store text chunks or embeddings in a non-parametric RAG-like database, potentially augmented with metadata for context \cite{mombaerts2024metaknowledgeretrievalaugmented}. More structured representations, such as GraphRAG, could also facilitate context-sensitive retrieval. However, capacity constraints on these types of databases may make it necessary to rely on more compressed parametric representations.

RQ2: \emph{How to segment continuous input into discrete episodes, and when to store them in an external memory?}

A major design question is \emph{when and how to segment} a continuous stream of agent experience into episodes to be encoded into an external memory. LLMs have already been shown to be capable of segmenting text into meaningful events, in a way that is similar to humans \cite{michelmann2023largelanguagemodelssegment}, and recent approaches show that further bundling related segments based on model surprise can improve long-term modeling \cite{behrouz2024titanslearningmemorizetest, fountas2024humanlikeepisodicmemoryinfinite}.

\emph{Leveraging long-context advances} can further improve encoding by providing a space in which new episodes can be equipped with a rich contextualization. Large hidden states or extended attention windows help capture high-fidelity contextual information, which can then be encoded into an external memory in a compressed format for future retrieval. 

\subsection{Retrieval}

RQ3: \emph{Given an external memory, how to select relevant past episodes for retrieval and reinstatement into in-context memory for the purpose of explicit reasoning?}

To employ past experiences in current tasks, an agent must \emph{retrieve} relevant episodes at the right time and \emph{reintegrate} them into its in-context memory with an adequate mechanism (Fig.\ref{fig:episodic-memory-llm-agent}, arrow (c)). Common strategies include prepending retrieved text tokens to the input sequence (as in RAG), manipulating representational states within the transformer (e.g., memory tokens \citep{bulatov2022recurrentmemorytransformer}), or adapting internal representations \citep{wu2024reftrepresentationfinetuninglanguage}.

RQ4: \emph{How can retrieval mechanisms in long-context LLMs improve and accelerate the optimization of external memory retrieval and reinstatement?}

\emph{Long-context advances} can be leveraged to inform when and what to retrieve at sequence lengths that are still feasible. Future research could explore tight integration of external memory with the model’s forward pass \citep{berges2024memorylayersscale} and adopt cross-architecture distillation \citep{wang2025mamballamadistillingaccelerating} to accelerate the development of external memory structures that retain many of the desirable properties of in-context memory while reducing the resource cost. 

\subsection{Consolidation}

RQ5: \emph{How to periodically consolidate external memory contents into the LLM's base parameters without forgetting previous knowledge?}

Eventually, merging external memory contents into the model’s parameters (Fig.\ref{fig:episodic-memory-llm-agent}, arrow (a)) promises to allow new generalized knowledge to be used without explicit retrieval. This process both prevents external memory overflow and supports continuous adaptation of the agent’s semantic and procedural backbone to the environment. Relevant techniques include context distillation, parametric knowledge editing, and localized fine-tuning methods that capture newly encountered information without catastrophic interference with other knowledge. Open questions remain about how to decide when to consolidate and how to compress many episodic instances into more abstract parametric knowledge while also retaining previous knowledge and skills.

\subsection{Benchmarks} 

RQ6: \emph{What new types of benchmarks are needed to assess episodic memory in LLM agents?}

Finally, \emph{evaluating} episodic memory effectiveness requires new tasks and metrics. Studies should test the recall of contextualized events after long delays, assessing how well agents remember when, where, and how events occurred. An example of such a study is the testing of instance-specific temporal order memory proposed by \citet{pink2024assessingepisodicmemoryllms}. Beyond controlled probes, benchmarks must incorporate real-world complexities: agents should demonstrate an improving task performance that is linked to encoding, retrieval, and consolidation of past experiences over extended timescales.

\section{Alternative Views}
\label{sec:alternatives}

While we argue that an explicit episodic memory framework is necessary for effective long-term and context-sensitive behavior, there are alternative perspectives suggesting that current or emerging methods might suffice in the future without the need for the concept of episodic memory to provide guidance.

\textbf{Scaling in-context memory will be sufficient.} One view suggests that advances in long-context methods—such as improved transformers, state-space models, or other architectures with extended context windows—will enable practically unlimited access to past information. Proponents claim that better positional encodings, modified attention mechanisms, and other in-context memory extensions will cover most relevant applications for LLM-based agents.

\textbf{Contextualized external memory will be sufficient.}
A second view holds that external memory structures—such as knowledge graphs or retrieval-augmented generation (RAG) systems—could eliminate the need for an episodic memory framework. By contextualizing data chunks and storing them in structured graphs, these systems aim to incorporate past context into current tasks effectively.

``Infinite'' in-context memory remains a speculative prospect. Extending limited context windows to include all information needed by an agent requires foreknowledge of the maximum timespan of relevant information. For very long timespans, this will either incur prohibitive computational costs or require compression methods that may lose key details. Only relying on external memory will still incur high storage costs, and require forgetting mechanisms. An episodic memory framework addresses these constraints by periodically consolidating information into high-capacity parametric memory (Figure \ref{fig:episodic-memory-llm-agent}, arrow (a)). This has the added benefit of enabling LLM agents to slowly improve over time, as they continue to learn from the past before they forget it.

\section{Conclusion}
This position paper argues that to fully realize efficient long-term LLM agents, we must endow LLM agents with episodic memory. We operationalize episodic memory---a term borrowed from cognitive science---for LLMs by highlighting five key characteristics that distinguish episodic memory from other types of memory in biological systems, and argue for why each property is also important for LLM agents. We position the call for episodic memory in LLM agents in the current literature and discuss how episodic memory can serve as a unifying goal for existing research directions. Lastly, we provide a roadmap of research questions towards implementing episodic memory in LLMs. By describing the potential of this research direction, we aim to spark a community-wide shift in how we conceive and engineer long-term memory in the move towards agentic AI---one that more deeply integrates lessons from cognitive science and brings together existing approaches in ML under a unifying goal with strong promise.

\nocite{langley00}

\bibliography{icml2025}
\bibliographystyle{icml2025}

\newpage
\appendix
\onecolumn

\end{document}